\documentclass{article}

\usepackage[preprint]{neurips_2020}

\usepackage[utf8]{inputenc} 
\usepackage[T1]{fontenc}    
\usepackage{hyperref}       
\usepackage{url}            
\usepackage{booktabs}       
\usepackage{amsfonts}       
\usepackage{nicefrac}       
\usepackage{microtype}      
\usepackage{bbm}
\usepackage{hyperref}

\usepackage[font=small,labelfont=bf]{caption}
\usepackage{subcaption}
\usepackage{todonotes}
\title{Deep Reinforcement Learning for Navigation in AAA Video Games}

\author{%
  Eloi Alonso\thanks{equal contribution} \\
  Ubisoft La Forge \\
  Montreal, Canada \\
  \texttt{eloi.alonso@ubisoft.com} \\
  \And 
  Maxim Peter$^*$ \\
  Ubisoft La Forge \\
  Montreal, Canada \\
  \texttt{maxim.peter@ubisoft.com} \\
  \And
  David Goumard \\
  Ubisoft La Forge \\
  Montreal, Canada \\
  \texttt{david.goumard@ubisoft.com} \\
  \And
  Joshua Romoff \\
  Ubisoft La Forge \\
  Montreal, Canada \\
  \texttt{joshua.romoff@ubisoft.com} \\
}

\begin{document}

\maketitle

\begin{abstract}
In video games, \textit{non-player characters} (NPCs) are used to enhance the players' experience in a variety of ways, e.g., as enemies, allies, or innocent bystanders. A crucial component of NPCs is navigation, which allows them to move from one point to another on the map. The most popular approach for NPC navigation in the video game industry is to use a \textit{navigation mesh} (NavMesh), which is a graph representation of the map, with nodes and edges indicating traversable areas. Unfortunately, complex navigation abilities that extend the character's capacity for movement, e.g., grappling hooks, jetpacks, teleportation, or double-jumps, increases the complexity of the NavMesh, making it intractable in many practical scenarios. Game designers are thus constrained to only add abilities that can be handled by a NavMesh if they want to have NPC navigation. As an alternative, we propose to use Deep Reinforcement Learning (Deep RL) to learn how to navigate 3D maps using any navigation ability. We test our approach on complex 3D environments in the Unity game engine that are notably an order of magnitude larger than maps typically used in the Deep RL literature. One of these maps is directly modeled after a Ubisoft AAA game. We find that our approach performs surprisingly well, achieving at least $90\%$ success rate on all tested scenarios. A video of our results is available at \url{https://youtu.be/WFIf9Wwlq8M}.

\end{abstract}

\section{Introduction}

Realistic navigation for \textit{non-player characters} (NPCs) is an important component in most video games to enhance the players' experience. The traditional pipeline for NPC navigation is as follows: 

\begin{enumerate}
  \item A graph representation of the world is pre-generated from the game geometry. 
  \item At runtime, a pathfinding algorithm like A*\citep{hart1968aFormalBasis} is applied on this graph to find the shortest path between any pair of locations in the game.
  \item A controller tailored to the character is used to follow this path. 
\end{enumerate}

The \textit{navigation mesh} (NavMesh) \citep{snook2000navmesh} is the most used representation of the world geometry \citep{macanlis_detailsNavMeshGen}. This graph, whose nodes represent the traversable surfaces of the 3D environment as convex polygons, is a compact representation of the world, independent of character abilities. Adding character constraints or abilities is traditionally done through other means, such as tweaking the pathfinding algorithm or extending the NavMesh with additional links (more details can be found in Section~\ref{subsec:BackgroundNavigation}). However, these approaches impose limitations on the kind of abilities that NPCs can use to navigate, which detract from the realism of the NPC.

In this paper, we set out to replace classical graph-based navigation with a system that can learn how to navigate between any two points on a map using all of the navigation options available to the character. Replacing the NavMesh with a learning system in modern AAA video games is challenging for several reasons. First, as the game worlds are increasingly more realistic, they have become both larger and more complex. Second, the maps can be dynamic as objects and other characters can move in the world. Finally, to replace the existing NavMesh, solutions need to run on a tight budget at runtime and ideally be relatively cheap to train.

To tackle these issues, we opt for a \textit{model-free} Reinforcement Learning (RL) approach \citep{sutton2018introduction} to the navigation problem, where an agent learns a \textit{policy} that maximizes a reward signal through interacting with an environment. Specifically, we train an agent with Deep RL to navigate to locations in the game world using Soft Actor-Critic (SAC) \citep{haarnoja2018soft} as our learning algorithm. We also build off of recent work that augments the agent's state with memory to effectively solve navigation tasks in complex $3$D environments \citep{mirowski2016learning, kapturowski2018recurrent, wijmans2019ddppo}. 

We begin by providing the relevant background and related work on navigation in Section~\ref{sec:Background}, followed by a detailed description of our system in Section~\ref{OurApproach}. Then, in Section~\ref{sec:experiments}, we demonstrate the performance of our Deep RL system, as well as several ablations, on two maps that we created using the Unity game engine \citep{juliani2018unity}. Notably, in contrast to previous RL-based approaches to navigation, we successfully train using continuous actions on 3D maps an order of magnitude bigger than simulators used in the research community \citep{wydmuch2018vizdoom, habitat19iccv}.

\section{Background and Related Work}
\label{sec:Background}

\subsection{Navigation in AAA games}
\label{subsec:BackgroundNavigation}

\paragraph{Waypoint graph:} Prior to the NavMesh, the most popular approach to game navigation was to use a \textit{waypoint graph} \citep{liden2002strategic}, which consists of a set of points of interest connected to each other. Unfortunately, this approach has several main drawbacks: its manual construction can be cumbersome and prone to human errors, it cannot handle dynamic objects, it is expensive to build since it needs to check all $n(n-1)$ combinations of paths, and the paths tend to not look realistic since all agents follow the same set of constrained paths \citep{tozour2002building}.  

\paragraph{NavMesh:} The NavMesh solves most of the aforementioned problems with waypoint graphs and is thus currently the main navigation tool used in video games \citep{macanlis_detailsNavMeshGen}. Specifically, the NavMesh divides the game map into a set of convex regions, which can each be trivially navigated within. It can be generated from either the raw geometry of the world using voxel-based approaches, or using pre-processed inputs like planar layers \citep{van2016ComparativeStudyNavMeshes, oliva2011AutomaticGenSuboptNavMesh}. It is thus a compact representation of the world's traversable terrains, independent of any character ability. Once the polygons have been placed, a graph is created by using the polygons as nodes and by connecting adjacent polygons with edges. With the built graph, search algorithms such as A* and Dijkstra's algorithm can be leveraged to find the shortest path between any two nodes. Then, the path is typically smoothed to look more realistic to the player \citep{brand2009efficient}. 

\paragraph{Using navigation abilities with a NavMesh:} 
\begin{figure}[!htbp]
    \centering
    \includegraphics[width=.8\textwidth]{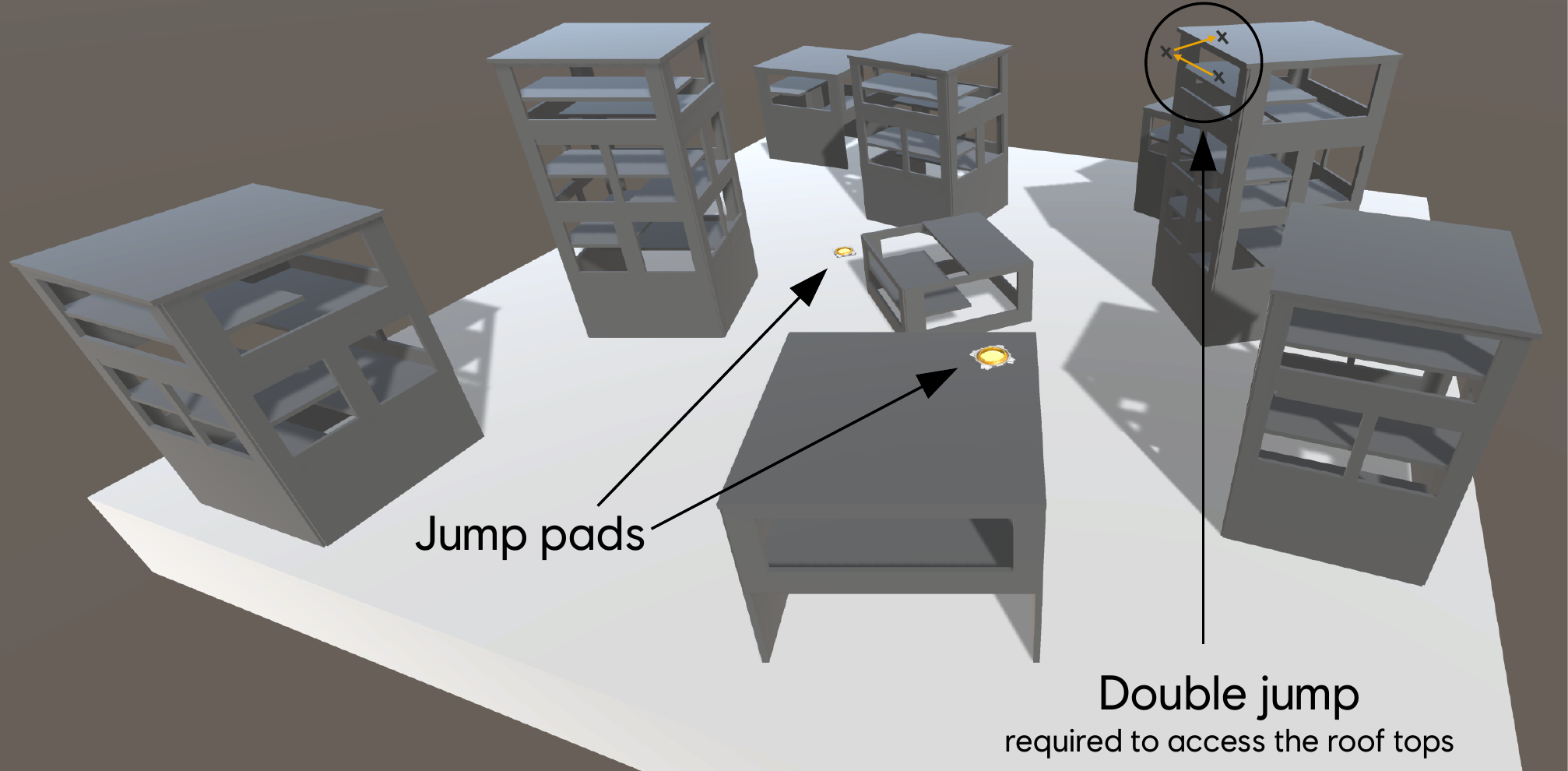}
    \caption{Overview of the Toy Map in Unity. The map is $120m \times 120m \times 30m$ with nine buildings of different heights. The navigation abilities of the agent are the following: move (forward / backward), strafe (left / right), rotate, jump (with the possibility to double jump), and use the two jump pads. Note that the rooftops have been purposely designed so that they are only accessible with the jump pads or with the double jump from the top floor (i.e., the agent must first jump outside and then jump again to reach the rooftop).}
    \label{fig:toymap}
\end{figure}
\begin{figure}[!htbp]
     \centering
     \begin{subfigure}[t]{0.49\textwidth}
         \centering
         \includegraphics[width=\textwidth]{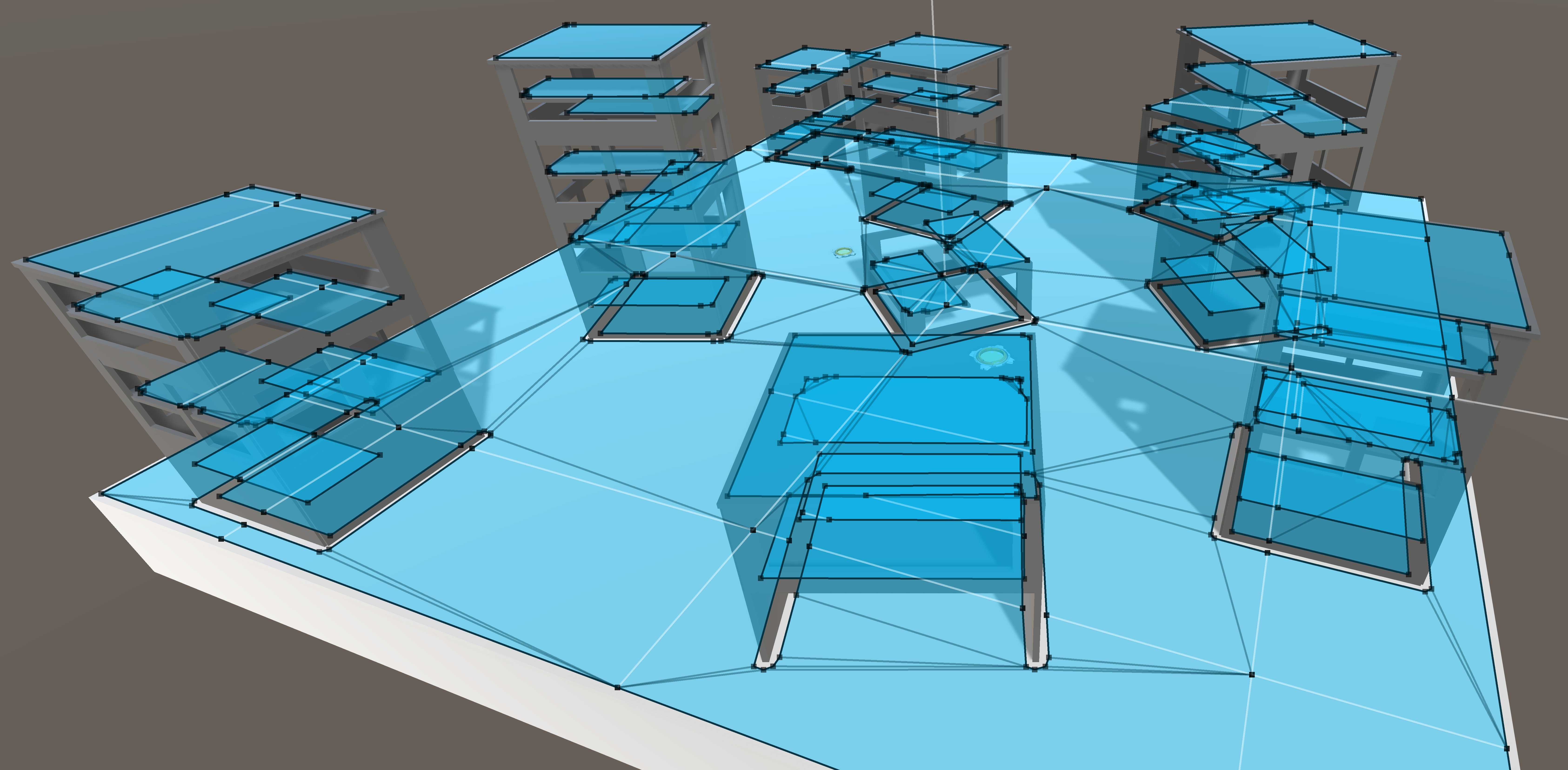}
         \caption{The NavMesh is automatically generated from the geometry. Each blue convex polygon is a node of the NavMesh and represents a traversable region of the map. NavMesh edges associate adjacent polygons. Here the NavMesh is not a connected graph since building floors and rooftops are not connected to anything.}
         \label{fig:toymap_navmesh_no_links}
     \end{subfigure}
     \hfill
     \begin{subfigure}[t]{0.49\textwidth}
         \centering
         \includegraphics[width=\textwidth]{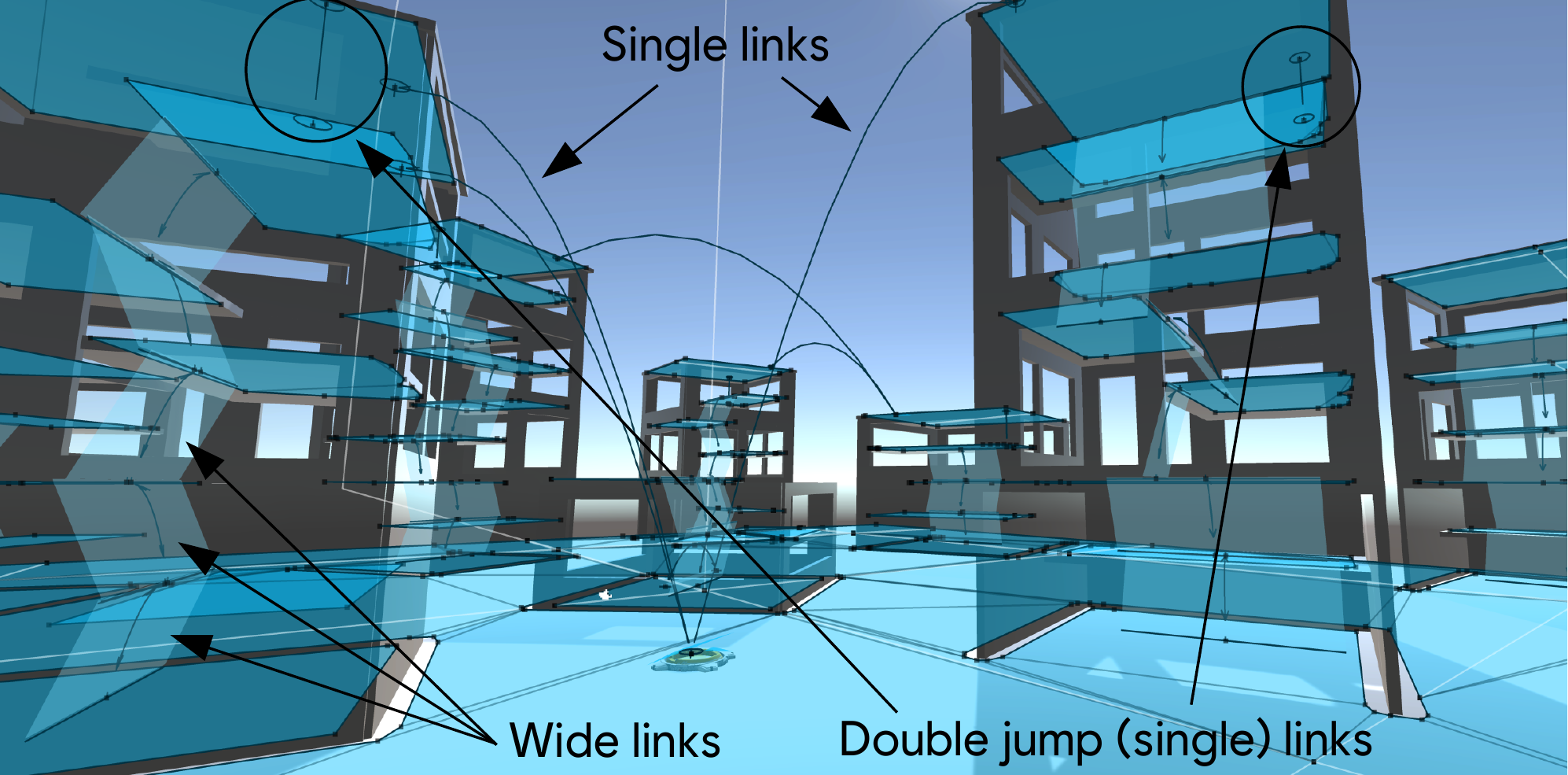}
         \caption{NavMesh links (either single or wide) are added to enable navigation between disconnected components of the graph. We added the minimum number of links such that the agent can reach any point on the map. For example, a link that connects the rooftop to its top floor is added to account for the rooftops being accessible via double jumps.}
         \label{fig:toymap_navmesh_with_links}
     \end{subfigure}
    \caption{NavMesh and additional links for the Toy Map.}
    \label{fig:toymap_navmesh}
\end{figure}

As alluded to in the introduction, it is increasingly common in modern games to offer additional navigation options to the players that enable the full use of the 3D space. For example, by using grappling hooks, jump-pads, jetpacks, teleportation, or other \textit{navigation abilities}, a player can very quickly navigate a map. These abilities add versatility to the players' gameplay but come at the cost of an increase in the number of feasible paths, making the NavMesh increasingly expensive to search through and labor intensive to create \citep{van2001quake, van2011navigation}. Specifically, in order to allow NPCs to use navigation abilities, the prevalent solution is to add an edge, called \textit{link}, connecting the nodes at both ends of the use of the ability. An animation is then played on the character as it goes through the link to give the illusion of using the ability.

A visual illustration of the NavMesh-based approach can be found in Figure~\ref{fig:toymap_navmesh}. We used Unity \citep{juliani2018unity} to build a minimal map, called \textit{Toy Map}, which can be seen in Figure~\ref{fig:toymap}. We then used Unity's built-in generation tool to create a NavMesh from the map geometry (Figure \ref{fig:toymap_navmesh_no_links}). We added a small set of complex navigation abilities (jumps, jump pads, and double jumps), and added some links so that the NavMesh can take these actions into account (see Figure~\ref{fig:toymap_navmesh_with_links}).

The search for all possible links that could be added to the graph can sometimes be done automatically. In the case of a jump, for example, possible jump trajectories are simulated and links are added between nodes that can be reached \citep{axelrod_graphGenerationInDynamicWorlds, Roumimper2017MeshNavThroughJumping, Budde2013AutoGenJumpLinks}. However, as this automatic search relies on connecting all the nodes that can be at the start and end of a movement trajectory, only the simplest trajectories can be used so that the search is tractable and the number of edges added is reasonable. 

In practice, having navigation abilities makes the number of links and the graph connectivity increase dramatically and results in extremely long computation time \citep{axelrod_graphGenerationInDynamicWorlds} with many redundant links \citep{Budde2013AutoGenJumpLinks}. In such cases, only adding a limited number of links manually is the favored option. However, not adding enough links can result in unrealistic behaviour, e.g., many NPCs aggregating in navigation bottlenecks around the map to use the specifically hand-designed way-points. As adding more links improves the realism of AI behaviors but increases the labor and runtime costs of the NavMesh, a compromise between efficiency and realism has to be found.

Another fundamental issue with links is that they do not correspond to a topological reality. For example, if a link to allow NPCs to climb a ladder is added, it is hard to interrupt the character once it starts climbing: Once on the link, the character is no longer moving in a traversable area, as the link does not correspond to a topological reality. Interrupting the character, e.g., by making him fall off the ladder, would require additional work to make it remain on the NavMesh.

Thus, existing solutions to support navigation abilities are far from perfect. Supporting navigation abilities comes at expensive labor and runtime costs, and fundamental flaws limit the realism of the obtained behaviors. In the case of our toy map, we manually placed a small number of additional links\footnote{The exact number of additional links is 54.} to bridge disconnected parts that should be accessible by using the jump, the double jump or the jump pads (Figure \ref{fig:toymap_navmesh_with_links}). While we are definitely not experts, adding links on such a small map still took us a few hours. Even if imperfect, this minimal example helps to illustrate the main limitations of the NavMesh and motivates our exploration of Deep RL for navigation in games.

\subsection{Navigation in Robotics and RL}

\paragraph{SLAM:} The classical approach to navigation in robotics is called \textit{simultaneous localization and mapping} (SLAM) \citep{leonard1991SLAM, durrant2006simultaneous}, which builds a high-level map (usually a top-down view) of the world from experience and locates the agent within this map. The high-level map is generated using data from sensors, such as LIDARs or RGB cameras. Once the map has been built, classical path-finding algorithms can be used to plan and extract the shortest paths between any two points. Recently, several works have extended the SLAM framework to learn the high-level mapping using differentiable neural networks \citep{gupta2017cognitive, zhang2017neural, beeching2020egomap} with some approaches even having success in video games \citep{bhatti2016playing}. Similar to the NavMesh, SLAM-based approaches struggle to integrate navigation abilities in the mapping.  Moreover, in the case of video games, we already have exact localization and mapping and thus do not need to use an estimate. 

\paragraph{Model-based RL:} Alternatively, in \textit{model-based} RL, a model of the \textit{transition dynamics}, i.e., a mapping from the current state and action of the agent to its next state, can be used for planning \citep{sutton1991dyna}. The model of the transition dynamics can either be estimated from interacting with the environment \citep{kaiser2019model}, or in some cases be given to the agent beforehand \citep{silver2016mastering}. In the case of video games, a model of the world can be used if we have access to the underlying game engine and the game state is resettable, in the sense that actions can be undone and different actions can be tried. 

Like SLAM and NavMesh-based approaches, model-based RL approaches could theoretically handle complex navigation actions and potentially be used to plan shortest paths. However, there are two notable drawbacks to using model-based approaches for planning. Firstly, like SLAM and NavMesh-based approaches, they are expensive to run at inference/mapping time as they need to compute many forward passes of the model to determine the best path using the complex navigation abilities. Secondly, when the model is estimated from data, they tend to suffer from compounding error due to model imperfections, which makes planning through the trained dynamics model challenging \citep{feinberg2018model, janner2019trust}.

\paragraph{Model-free RL:} \textit{Model-free} RL does not use a model to plan but learns which actions to take in the environment through pure trial and error \citep{sutton2018introduction}. Previous works have used model-free RL for navigation \citep{mirowski2016learning, wijmans2019ddppo} but have been mostly limited to relatively small 2D environments with simple action spaces. Recent works have circumvented the lack of planning in model-free RL by using a hierarchical architecture, where intermediate goals are given to a controller by a high level planner \citep{bansal2020combining, eysenbach2019search, meng2019scaling}. As navigation in a visually complex environment is usually modeled as a \textit{partially observable markov decision process}, the importance of using memory has been previously acknowledged \citep{mirowski2016learning}. While unstructured memory such as LSTMs \citep{hochreiter1997long} can be used, architectures involving spatially structured memory have also been explored \citep{parisotto2017neural, beeching2020egomap}. The use of auxiliary tasks to accelerate the learning of challenging goal-based RL problems has also been a subject of study \citep{mirowski2016learning, andrychowicz2017hindsight, ghosh2019learning}.

\section{Approach}
\label{OurApproach}

The following section describes our approach to solve point-to-point navigation on a fixed 3D map using navigation abilities available to the agent. 

\begin{figure}[!htbp]
    \includegraphics[width=\textwidth]{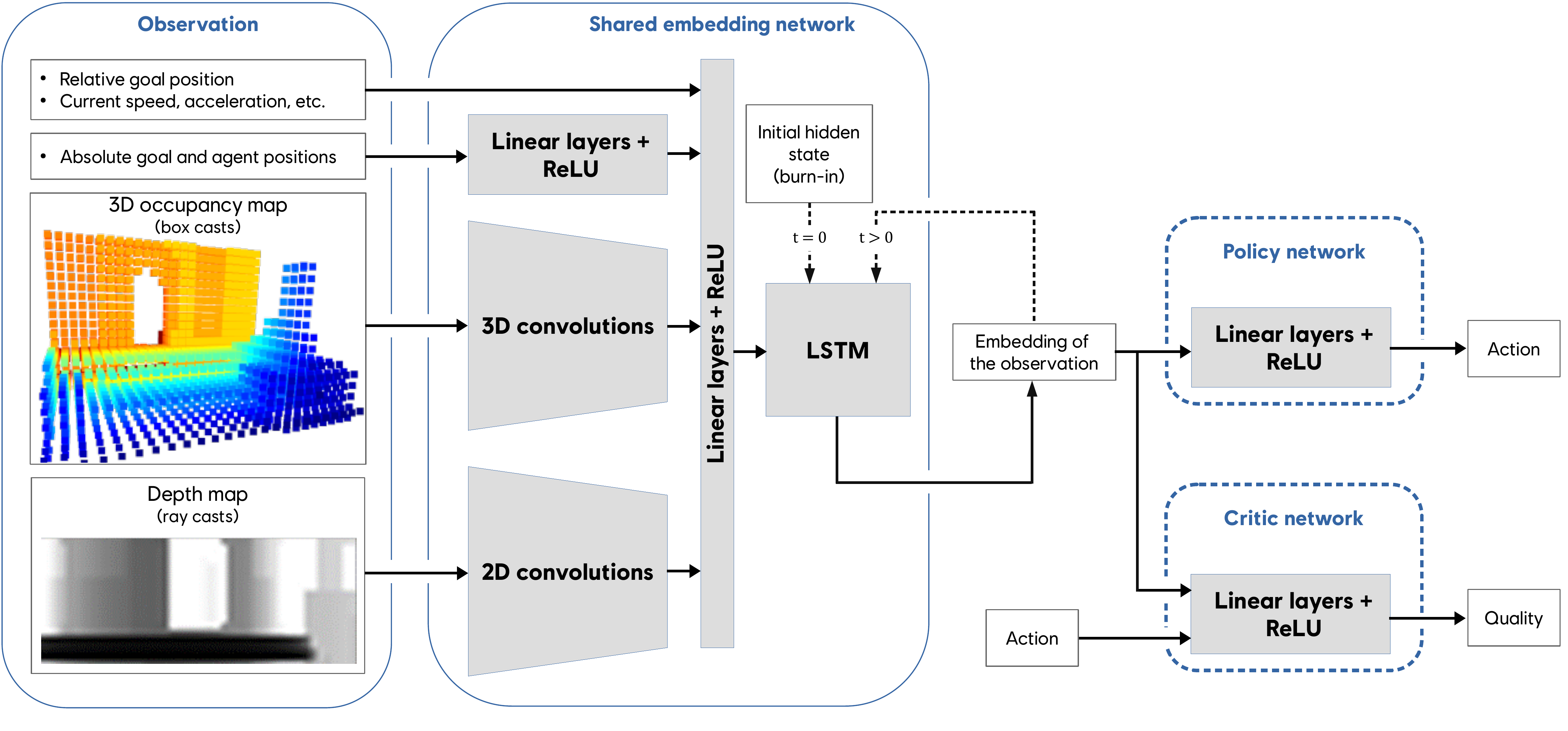}
    \caption{Architecture of our system.  The $3$D occupancy map, $2$D depth map, and absolute goal and agent positions, pass through independent feature extraction layers ($3$D convolutions, $2$D convolutions, and linear layers respectively). The output of each feature extractor is then combined with other state variables, such as relative goal position, speed, acceleration, and previous action. The combined output is fed through several linear layers, followed by an LSTM, to create the final embedding shared by both the policy and critic heads (trained using the critic loss).}
    \label{fig:architecture}
\end{figure}

\paragraph{States:} The state is composed of local perception in the form of a $3$D occupancy map and a $2$D depth map, as well as scalar information about physical attributes of the agent and its goal (velocity, relative goal position, absolute goal position, and previous action). The $3$D occupancy map, referred to as \textit{BoxCasts}, can be generated and cached offline. This makes it efficient to compute at runtime which is critical when running inside a video game engine. $2$D depth maps have been used in several recent works \citep{gupta2017cognitive, bansal2020combining, habitat19iccv, wijmans2019ddppo}. It can be extracted from a rendering camera, or in our case by casting \textit{RayCasts}. The absolute positions of the agent and its goal pass through their own network to extract an absolute position embedding similarly to the goal encoder in \citep{wijmans2019ddppo}. Notably, by providing the position of the agent the problem is no longer partially observable because the map is static.

\paragraph{Actions:} The actions are continuous, and correspond to \textit{jump, forward, strafe, rotate}. The jump is treated as a continuous action on the algorithmic side and binarized in the environment. 

\paragraph{Rewards:} To avoid complications associated with long term credit assignment when using a sparse reward, we densify the reward signal to be $R_t = \max(\min_{\forall i \in [|0, t-1|]} D_i(agent, goal) - D_t(agent, goal), 0) + \alpha + \mathbbm{1}_{D_t(agent, goal) \leq \epsilon}$ where $D_t$ is the Euclidean distance between the positions of its arguments at time $t$, $\alpha$ is a penalty given at each step and $\epsilon$ is the distance below which the agent is considered to have reached its goal. Intuitively, this reward signal encourages the agent to get closer to its goal and reach it as fast as possible.

\paragraph{Training procedure:} As running a game engine is costly and in order to be more sample efficient, we use an off-policy RL algorithm called Soft Actor-Critic \citep{haarnoja2018soft} modified so that the entropy coefficient is learned and there is no state value network \citep{haarnoja2018SACApplications}. The critic and policy networks share layers that are tasked with extracting an embedding from local perception as well as previous steps by using Convolutional Neural Networks \citep{lecun1989backpropagation} and an LSTM \citep{hochreiter1997long,Gers2000}. Furthermore, to train an off-policy RL algorithm with an LSTM, we use a burn-in to initialize the hidden states as recommended in \citep{kapturowski2018recurrent, R2D3}.

During training, we spawn the agent and its goal in a cylinder with a variable radius. An episode is considered over when the agent has reached its goal or when the number of steps is over a certain budget. To allow the agent to have informative trajectories at all stages during training, we use a training curriculum \citep{bengio2009curriculum} and increase the radius of the spawning cylinder until the full map is covered. All the networks are trained using the Adam optimizer \citep{kingma2014adam}. More details on our hyperparameters can be found in Table~\ref{hyperparameters} and  Figure~\ref{fig:architecture} describes our architecture.

\section{Experiments}
\label{sec:experiments}

\begin{figure}[!htbp]
    \centering
    \includegraphics[width=\textwidth]{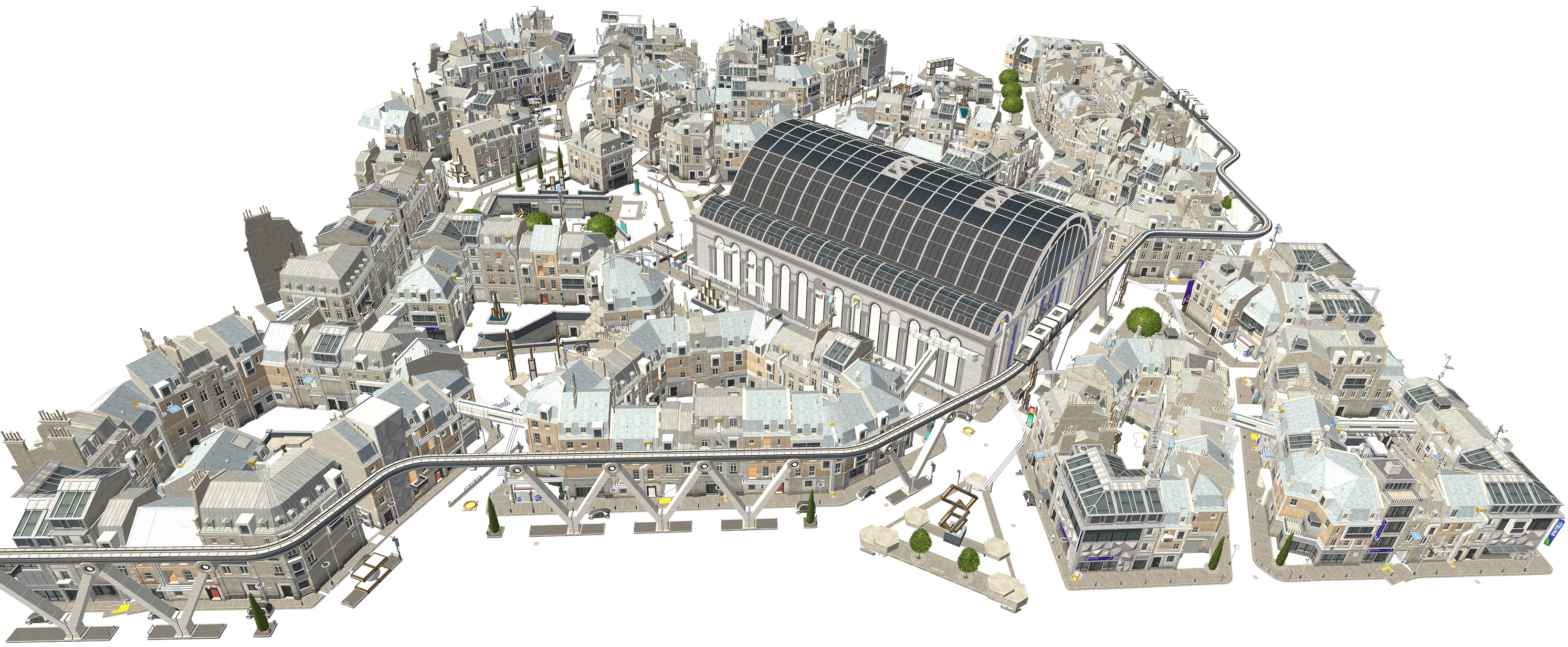}
    \caption{Overview of the Big Map in Unity. The map is $300m \times 300m \times 100m$ and features complex environments with a lot of buildings, making it far more realistic than our Toy Map. Due to the goals being placed all over the map, e.g., on top of buildings, the usage of double jumps and jump pads is required.}
    \label{fig:bigmap}
\end{figure}

\begin{figure}
     \centering
         \includegraphics[width=0.45\textwidth]{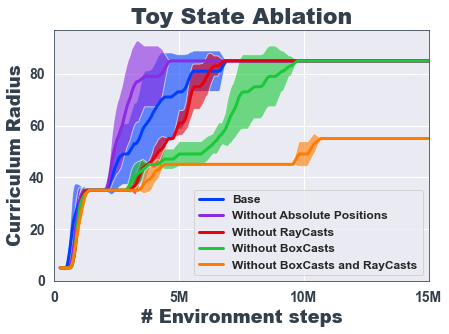}
         \includegraphics[width=0.45\textwidth]{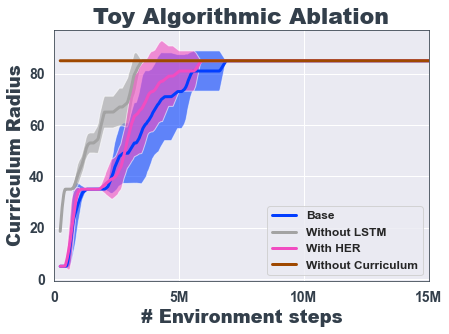}
        
        \includegraphics[width=0.45\textwidth]{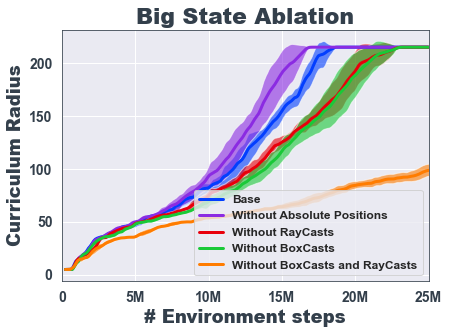}
          \includegraphics[width=0.45\textwidth]{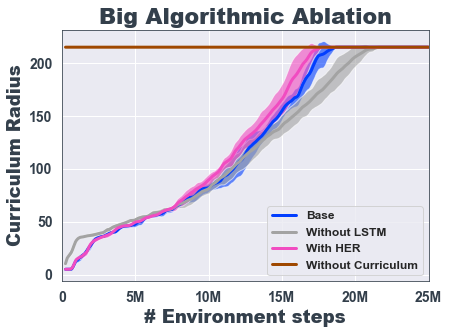}
    \caption{All agents were trained over $5$ seeds. The shaded regions represent $95\%$ confidence intervals. We run a state ablation (left) and algorithmic ablation (right) on the Toy Map (top) and the Big Map (bottom). The y-axis is the current radius of the curriculum. The maximum radius is $85$ for the Toy Map and $215$ on the Big Map.}
    \label{fig:ablation_line}
\end{figure}

\begin{figure}
     \centering
         \includegraphics[ width=0.49\textwidth]{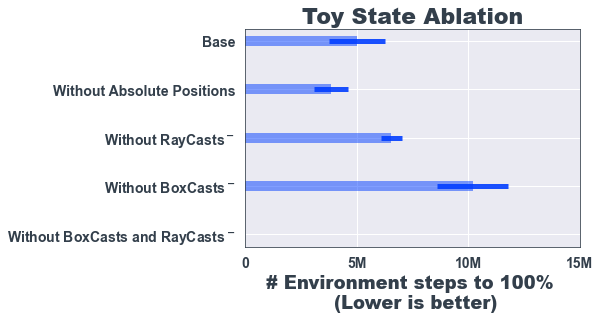}
         \centering
         \includegraphics[width= 0.41 \textwidth]{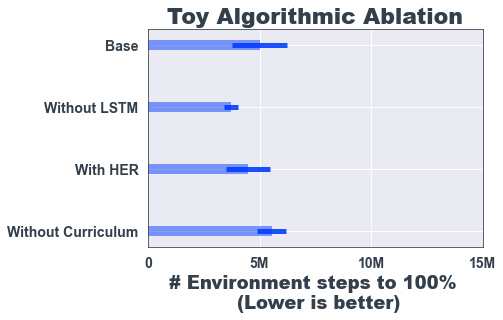}
         
         \includegraphics[ width=0.49\textwidth]{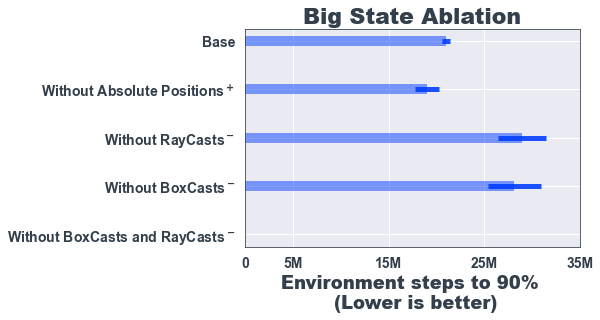}
         \centering
         \includegraphics[width= 0.41 \textwidth]{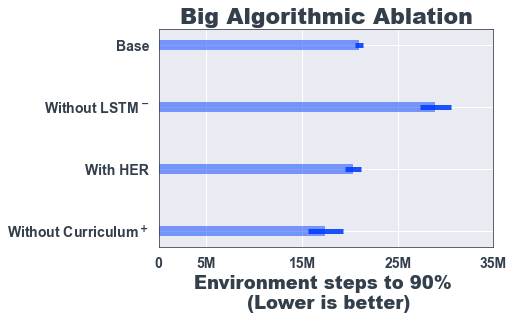}
         
         \caption{All agents were trained over $5$ seeds. The error bars represent $95\%$ confidence intervals. We run a state ablation (left) and an algorithmic ablation (right) on the Toy Map (top) and Big Map (bottom). The x-axis measures the number of environment steps that it takes the given agent to achieve the target success rate ($100\%$ on the Toy Map and $90\%$ on the Big Map) on the final curriculum level. We note that the ablation without a curriculum starts at the final level. We run significance testing using Welch’s t-test as recommended in the Deep RL literature \citet{colas2019hitchhiker,henderson2018deep}. $^-$ indicates the method was significantly ($p<0.05$) worse than \textit{Base}, $^+$ indicates a significant improvement over \textit{Base}. }
         
    \label{fig:ablation_bar}
\end{figure}

To highlight the capabilities of our Deep RL system, we train an agent as described in Section~\ref{OurApproach} on both a Toy Map (see Figure \ref{fig:toymap}) and a Big Map (see Figure \ref{fig:bigmap}), which we built using the Unity game engine \citep{juliani2018unity}. In order to determine the factors that lead to its success, we also run two separate sets of ablations of our method on both maps: state-based and algorithmic ablations. For the state-based ablation, we evaluate our agent without BoxCasts, without RayCasts, without BoxCasts and RayCasts, and without absolute positions. For the algorithmic ablation study, we compare our Base agent against agents that were trained without an LSTM, without a curriculum, and using Hierarchical Experience Replay (HER) \citep{andrychowicz2017hindsight}.
\paragraph{Ablation results:} 
The results for both ablation studies can be found in Figure~\ref{fig:ablation_line} and Figure~\ref{fig:ablation_bar}. Concerning the state ablation, we find that, on both the Toy Map and the Big Map, the Base agent is significantly faster (in terms of samples) at reaching the target success rate on the final curriculum level than without BoxCasts, without RayCasts, and without BoxCasts and RayCasts (see Figure~\ref{fig:ablation_bar}). Furthermore, we find that removing the absolute position from the agent's state does not negatively impact performance. In fact, the agent performs slightly better on average (over $5$ seeds) which we find to be statistically significant ($p<0.05$) on the Big Map but not significant on the Toy Map. 

Moreover, we find that by removing both BoxCasts and RayCasts, the agent fails to reach the final curriculum level in the training period (see Figure~\ref{fig:ablation_line}) on both the Toy Map and the Big Map. As this agent is training without local perception, it is perhaps unsurprising that it is less sample efficient.  However, we do note, that since the performance of the agent is still increasing (albeit slowly see Figure~\ref{fig:ablation_line}), given more training samples it could be possible that it attains the target success rate ($100\%$ on the Toy Map and $90\%$ on the Big Map) on the final curriculum level. To verify this, we ran that configuration for $100M$ steps on the Big Map and confirmed that it could not reach the target success rate of $90\%$ (see Figure~\ref{fig:ablation_line_longer_appendix} in the appendix), but plateaus at about $80\%$ success rate on the final curriculum level. 

In terms of the algorithmic ablation, none of the algorithmic changes make a significant difference on the Toy Map (see Figure~\ref{fig:ablation_bar}). On the other hand, on the Big Map we find that removing the LSTM significantly hurts sample efficiency.  We hypothesize that the discrepancy between the Toy Map and Big Map with regards to the LSTM can be explained by the simplicity of the Toy map, with the local perception being sufficient to accurately represent the space. We also find that removing the training curriculum improves performance significantly on the Big Map, whereas it does not have a significant impact on the Toy Map. Upon further inspection, we found that we increased the curriculum radius by $5$ meters at every iteration on the Big Map and by $10$ meters on the Toy Map. A hyperparameter search would need to be done in order to find the optimal setting. However, we note that the strong performance without a curriculum on the Big Map indicates that the dense reward signal that we use is sufficient. Finally, we find that performance does not increase significantly with the use of Hierarchical Experience Replay (HER) \citep{andrychowicz2017hindsight} on the Toy Map but it does (slightly) on the Big Map. Since HER has been shown to be effective in sparse reward regimes, the minimal gains could be attributed to our use of a dense reward signal during training.

\paragraph{Deep RL vs. NavMesh:}
\label{par:DeepRlVSNavMesh}
To provide a comparison to the NavMesh, we made available a video\footnote{\url{https://youtu.be/WFIf9Wwlq8M}} that shows how the two systems behave on the Toy Map. As explained in Section~\ref{subsec:BackgroundNavigation}, such a comparison is imperfect as the NavMesh-based navigation could be made better by spending more time adding links and animating characters on link traversals. Nevertheless, we believe that this visual comparison provides interesting insights to understand how our approach relates to the classical solution in terms of using navigation abilities.

\section{Discussion}

The results of our ablation study indicate that local perception (through both RayCasts and BoxCasts) drastically improves the sample efficiency of the Deep RL system. We also found that the LSTM was a crucial component for optimal sample efficiency when training on large maps. To further improve sample efficiency, in future work, auxiliary tasks \citep{mirowski2016learning, jaderberg2017unreal, ghosh2019learning} could be used to help train the representation layers more efficiently. 

All of the experiments conducted in this paper were concerned with the sample efficiency of the Deep RL agent on a single map. We emphasize that sample efficiency is indeed important when faced with a \textit{slow} simulator, which is typically the case with AAA games. While this may be alleviated by specifically redesigning the game engine into a simple binary that can be run much faster than real-time \citep{vinyals2019grandmaster}, this may not always be feasible due to engineering complications.
Sample efficient solutions, however, often come at the cost of increased inference time. As we operate in video games, we need a solution that is both sample efficient to train and that can be evaluated at runtime. We run our solution on internal tools for Neural Network inference in game engines which make it viable to be used in games.

Finally, while not addressed in this paper, generalization is an interesting direction of future work for navigation in AAA games. Typically the agent is tasked with navigation within a game with dynamic components, e.g., players or objects. As for the NavMesh, it is usually prebuilt to optimize for runtime efficiency and thus cannot easily handle dynamic components \citep{macanlis_detailsNavMeshGen}. 
Instead, NavMeshes are either updated dynamically at runtime \citep{marden_dynamicUpdateNavMesh} or navigate around non-static objects by using heuristics \citep{brand2009efficient}. Thus, a generalizable Deep RL agent may be key to adapt to dynamic components that may not have appeared identically during training.  Generalization could also enable the reusability of the trained system on novel maps. This can have drastic cost saving impacts when the number of maps becomes large, e.g. when procedurally generated, where retraining the agent is simply not feasible. 

In this paper, we showed that Deep RL can be an alternative to the NavMesh for navigation in complicated 3D maps, such as the ones found in AAA video games. In comparison with previous works exploring Deep RL for navigation, our Big Map is an order of magnitude bigger \citep{habitat19iccv, wydmuch2018vizdoom}. Unlike the NavMesh, the Deep RL system is able to handle navigation actions without the need to manually specify each individual link. We also performed a thorough ablation study to determine the main factors that lead to the success of the Deep RL approach. We find that our approach performs surprisingly well, achieving at least $90\%$ success rate on all tested scenarios.

\section*{Acknowledgments}

We would like to thank Julien L'Heureux, Claudine Combe, Nicolas Landron, Bérenger Bailly, Mike Yurick, Philippe Marcotte and Adrien Logut for many engineering contributions and thorough feedback during the development of this project. We also wish to thank Pierre Falticska, Olivier Pomarez, Paul Barde and Julien Roy for insightful discussions and their feedback on an earlier draft of this paper. Finally, we thank Olivier Delalleau, Batu Aytemiz and Sahand Rezaei-Shoshtari for their contribution to an early iteration of this project.

\bibliography{neurips_2020}

\newpage
\section{Appendix}

\begin{table}[!htbp]
\centering
 \begin{tabular}{c | c} 
 \hline
  Parameter & Value \\ 
  \hline
  Optimizer & Adam \\
  Learning rate & $3 \times 10^{-4}$ \\
  Discount ($\gamma$) & $0.99$ \\
  Replay buffer size & $4000000$ \\
  Initial alpha & $0.05$ \\
  Neurons per hidden layer & $1024$ \\
  Number of hidden layers & $2$ \\
  Neurons per LSTM layer & $1024$ \\
  Number of LSTM layers & $1$ \\
  Burn in & $20$ \\
  LSTM sequence length & $60$ \\
  Activation function & ReLU \\
  Minibatch size & $64$ \\
  Target smoothing coefficient & $0.005$ \\
  Training / environment steps& $4$ \\
  Curriculum increase condition & Success rate > 80\%\\
  \hline
\end{tabular}
\caption{Our hyperparameters}
\label{hyperparameters}
\end{table}

\begin{figure}
     \centering
         \includegraphics[width=\textwidth]{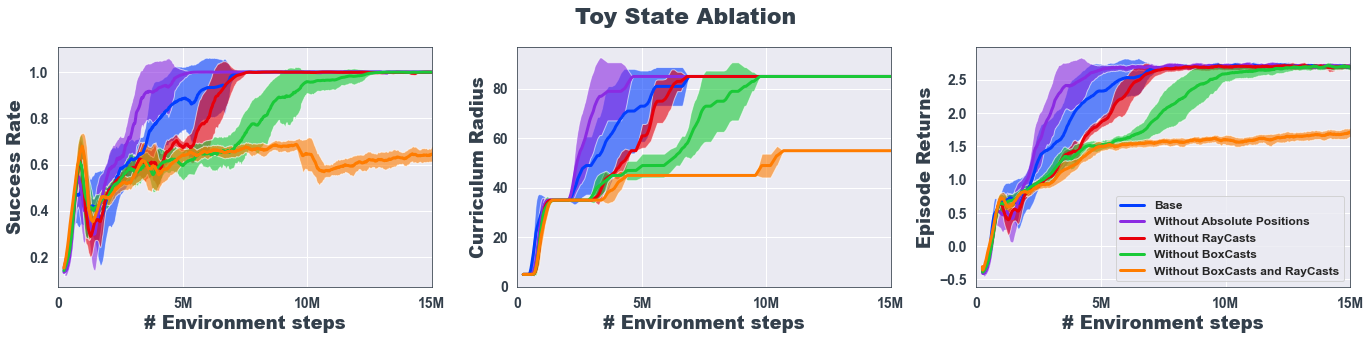}

         \includegraphics[width=\textwidth]{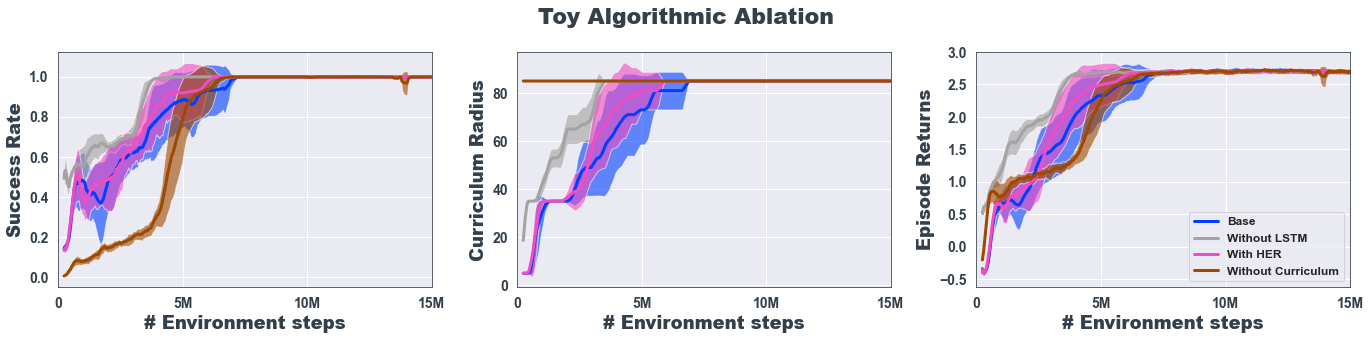}
         
          \includegraphics[width=\textwidth]{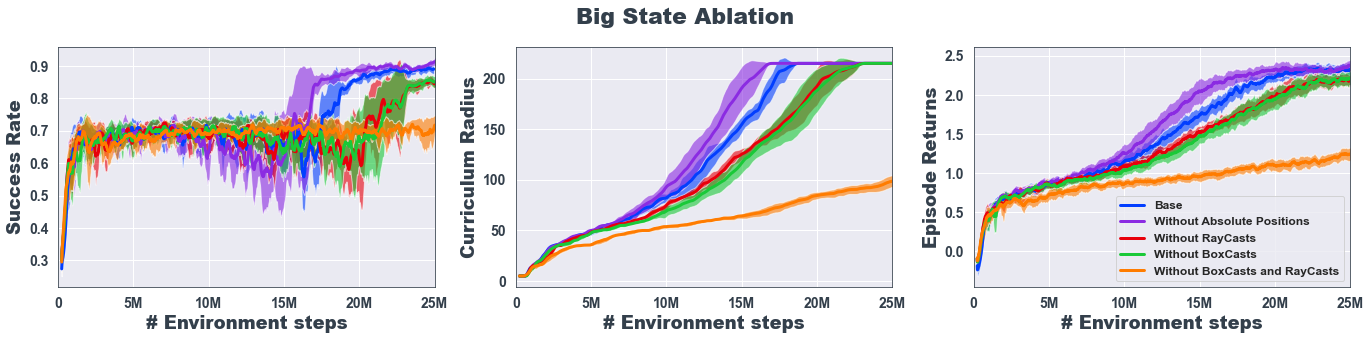}

         \includegraphics[width=\textwidth]{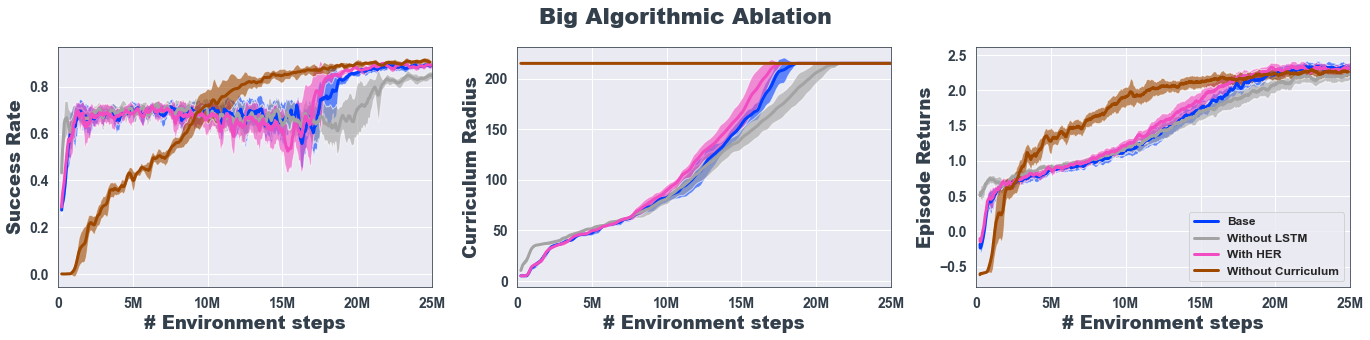}
    \caption{All agents were trained over $5$ seeds. The shaded regions represent $95\%$ confidence intervals. We run a state ablation and algorithmic ablation on the Toy Map (top $2$) and Big Map (bottom $2$). (left) The success rate on the current curriculum (1 indicates 100\% success rate). (middle) The current radius of the curriculum (the maximum radius is $85$ for the Toy Map and $215$ for the Big Map). (right) The return of the agent given the current curriculum. We note that the maximum return changes based on the curriculum, thus, it is best to compare algorithms at the same curriculum radius.}
    \label{fig:ablation_line_appendix}
\end{figure}

\begin{figure}
     \centering
\includegraphics[width=\textwidth]{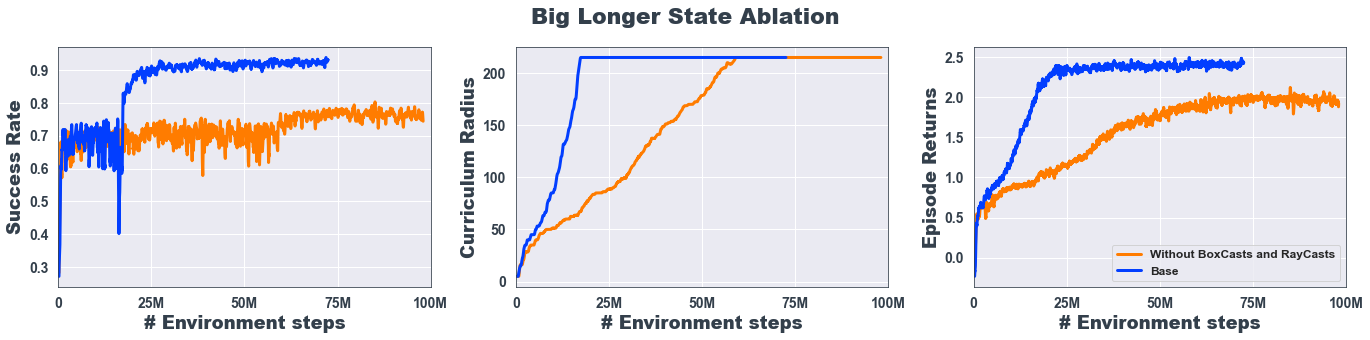}

    \caption{Longer training run for the Base algorithm and without BoxCasts and RayCasts. The success rate on the current curriculum (1 indicates 100\% success rate). (middle) The current radius of the curriculum (the maximum radius is $85$ for the Toy Map and $215$ for the Big Map). (right) The return of the agent given the current curriculum. We note that the maximum return changes based on the curriculum, thus, it is best to compare algorithms at the same curriculum radius.}
    \label{fig:ablation_line_longer_appendix}
\end{figure}
\end{document}